%% file: main.tex
\documentclass{article}

     \usepackage[preprint]{neurips_2019}

\usepackage[utf8]{inputenc} % allow utf-8 input
\usepackage[T1]{fontenc}    % use 8-bit T1 fonts
\usepackage{hyperref}       % hyperlinks
\usepackage{url}            % simple URL typesetting
\usepackage{booktabs}       % professional-quality tables
\usepackage{amsfonts}       % blackboard math symbols
\usepackage{nicefrac}       % compact symbols for 1/2, etc.
\usepackage{microtype}      % microtypography

\usepackage{subfiles}
\usepackage[normalem]{ulem}
\usepackage{microtype}
\usepackage{url}
\usepackage{subfigure}
\usepackage{booktabs}
\usepackage{amsfonts,xcolor,amsthm,bbm,mathrsfs,amsmath}
\usepackage{algorithm,algorithmic,bm,color}
\usepackage{multirow}
\usepackage{wrapfig}
\usepackage{adjustbox}
\usepackage{graphics, color,  dsfont}
\usepackage[nottoc]{tocbibind}

\usepackage{natbib}
\setcitestyle{square, comma}
\def\comment#1{}

\newcommand{\ts}{^{\top}}
\newcommand{\ie}{{i.e.}}
\newcommand{\eg}{{e.g.}}

\title{CDSA: Cross-Dimensional Self-Attention for Multivariate, Geo-tagged Time Series Imputation}
\author{%
\normalfont Jiawei Ma$^1$\thanks{indicates equal contributions.}
\and
Zheng Shou$^{1*}$
\and
Alireza Zareian$^1$
\and 
Hassan Mansour$^2$
\and
Anthony Vetro$^2$
\and
Shih-Fu Chang$^1$
\and
\\
$^{1}$Columbia University \hspace{30pt} $^{2}$Mitsubishi Electric \hspace{20pt} 
}

\begin{document}

\maketitle
%%%%%%%%%%%%%%%%%%%%%%%%%%%%%%%%%%%%%%%%%%%%%%%%%%%%%%%%%%%%%%%%%%%%%%%%%%%%%%%%%%
%%%%%%%%%%%%%%==================     Abstract      ==================%%%%%%%%%%%%%
%%%%%%%%%%%%%%%%%%%%%%%%%%%%%%%%%%%%%%%%%%%%%%%%%%%%%%%%%%%%%%%%%%%%%%%%%%%%%%%%%%
\begin{abstract}

Many real-world applications involve multivariate, geo-tagged time series data: at each location, multiple sensors record corresponding measurements.
For example, air quality monitoring system records PM2.5, CO, etc.
The resulting time-series data often has missing values due to device outages or communication errors. 
In order to impute the missing values, state-of-the-art methods are built on Recurrent Neural Networks (RNN), which process each time stamp sequentially, prohibiting the direct modeling of the relationship between distant time stamps. 
%Recently, a self-attention mechanism has been proposed for sequence modeling tasks such as machine translation, significantly outperforming RNNs since the temporal relationship between measurements can be modeled explicitly.
%Such time series data often have missing values due to device damages or communication errors.
%In order to impute the missing values, the state-of-the-art methods have been built on Recurrent Neural Networks (RNN) which has to process each time stamp sequentially, prohibiting the direct modeling of the relationship between each two time stamps.
Recently, the self-attention mechanism has been proposed for sequence modeling tasks such as machine translation, significantly outperforming RNN because the relationship between each two time stamps can be modeled explicitly.
%However, the relevance across different dimensions  has not been fully explored.
%Given the success of self-attention mechanism in modeling long-term dependencies between words in sentence,
In this paper, we are the first to adapt the self-attention mechanism for multivariate, geo-tagged time series data.
In order to jointly capture the self-attention across multiple dimensions, including time, location and the sensor measurements, while maintain low computational complexity, we propose a novel approach called Cross-Dimensional Self-Attention (CDSA) to process each dimension sequentially, yet in an order-independent manner. 
Our extensive experiments on four real-world datasets, including three standard benchmarks and our newly collected NYC-traffic dataset, demonstrate that our approach outperforms the state-of-the-art imputation and forecasting methods. 
A detailed systematic analysis confirms the effectiveness of our design choices.
\end{abstract}

\subfile{Paper_Intro}
\subfile{Paper_Model}
\subfile{Paper_Result}
%%%%%%%%%%%%%%%%%%%%%%%%%%%%%%%%%%%%%%%%%%%%%%%%%%%%%%%%%%%%%%%%%%%%%%%%%%%%%%%%%%
%%%%%%%%%%%%%%================      Conclusion       ================%%%%%%%%%%%%%
%%%%%%%%%%%%%%%%%%%%%%%%%%%%%%%%%%%%%%%%%%%%%%%%%%%%%%%%%%%%%%%%%%%%%%%%%%%%%%%%%%
\section{Conclusion}

In this paper, we have proposed a cross-dimensional self-attention mechanism to impute the missing values in multivariate, geo-tagged time series data.
We have proposed and investigated three methods to model the crossing-dimensional self-attention.
%Instead of imposing the assumptions over the data generating process, we treat the missing value as variables of the \textit{Encoder} so that the \textit{altered loss} is obtained without limitation from time sequence, identity network and measurement distribution to make the estimation of missing value much more accurate. 
%We fit the incomplete data into a \textit{Transformer} framework for prediction while the \textit{Encoder} still performs as a imputation \& prediction manner and obviously decrease the discrepancy for long-time prediction.
Experiments show that our proposed model achieves superior results to the state-of-the-art methods on both imputation and forecasting tasks.
Given the encouraging results, we plan to extend our CDSA mechanism from multivariate, geo-tagged time series to the input that has higher dimension and involves multiple data modalities.
Furthermore, we will publicly release the collected NYC-traffic dataset for future research.
%Furthermore, we will publicly release the collected NYC-traffic dataset for future research.
%In the future, it would be interesting to extend our CDSA crossing three dimensions to multiple dim
%\az{\sout{adapted self-attention mechanism from NLP} 

\section{Acknowledgments}

This work is supported by Mitsubishi Electric.

%\newpage
%\subfile{Section/Paper_Supp}

\medskip

{\small
\bibliography{Section/reference}
}

\end{document}

%% file: Paper_Intro.tex
%%%%%%%%%%%%%%%%%%%%%%%%%%%%%%%%%%%%%%%%%%%%%%%%%%%%%%%%%%%%%%%%%%%%%%%%%%%%%%%%%%
%%%%%%%%%%%%%%==================   Introduction    ==================%%%%%%%%%%%%%
%%%%%%%%%%%%%%%%%%%%%%%%%%%%%%%%%%%%%%%%%%%%%%%%%%%%%%%%%%%%%%%%%%%%%%%%%%%%%%%%%%
\section{Introduction}

Various monitoring applications, such as those for air quality~\cite{zheng2015forecasting}, health-care~\cite{silva2012predicting} and traffic~\cite{jagadish2014big}, widely use networked observation stations to record multivariate, geo-tagged time series data.
For example, air quality monitoring systems employ a collection of observation stations at different \textbf{locations}; at each location, multiple sensors concurrently record different \textbf{measurements} such as PM2.5 and CO over \textbf{time}.
Such time series are important for advanced investigation and useful for many downstream tasks including classification, regression, forecasting, etc.
However, due to unexpected sensor damages or communication errors, missing data is unavoidable.
It is very challenging to impute the missing data because of the diversity of the missing patterns: sometimes almost random while sometimes following various characteristics.
%; the missing rate is sometimes scarce while sometimes severe.
%intractable
%\textcolor{red}{[The malfunction of each sensor induces missing value individually]}, leading to various missing patterns.
%\zs{list two pattern examples?}
%Meanwhile, the system error will cause missing values cross all possible locations and measurements at the same time.

\begin{figure}[t!]
    \centering
    \includegraphics[width=\linewidth]{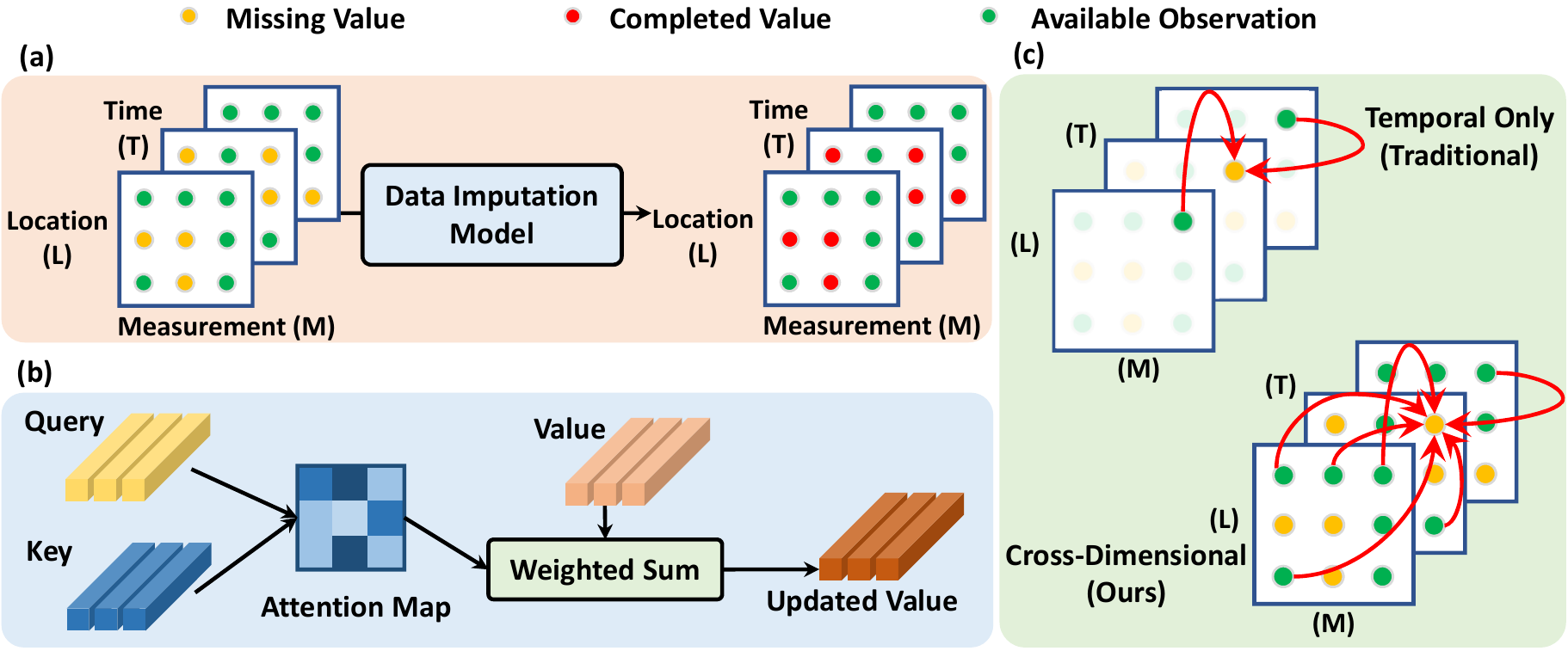}
    \caption{
    (a) \textbf{Illustration of the multivariate, geo-tagged time series imputation task}: the input data has three dimensions (i.e. time, location, measurement) with some missing values (indicated by the \textcolor{orange}{orange} dot); the output is of same shape as the input while the missing values have been imputed (indicated by the \textcolor{red}{red} dot).
    (b) \textbf{Self-attention mechanism}: the \textbf{Attention Map} is first computed using every pair of \textbf{Query} vector and \textbf{Key} vector and then guides the updating of \textbf{Value} vectors via weighted sum to take into account contextual information.
    (c) \textbf{Traditional Self-Attention} mechanism updates Value vector along the temporal dimension only vs. \textbf{Cross-Dimensional Self-Attention} mechanism updates Value vector according to data across all dimensions.
    }
    \label{Fig:Motivation}
\end{figure}

Traditional data imputation work usually suffer from imposing strong statistical assumptions. % and may fail to model the measurement correlation. 
For example, \cite{scharf1991statistical,friedman2001elements} generally tries to fit a \emph{smooth curve} on observations in either time series~\cite{ansley1984estimation,shumway1982approach} or spatial distribution ~\cite{friedman2001elements,stein2012interpolation}.
%, \eg, ARIMA~\cite{ansley1984estimation} and EM algorithm~\cite{shumway1982approach}, 
%, \eg, KNN~\cite{friedman2001elements} and Kriging Interpolation~\cite{stein2012interpolation}.
Deep learning methods~\cite{li2018dcrnn_traffic,che2018recurrent,cao2018brits,luo2018multivariate} have been proposed to capture temporal relationship based on RNN ~\cite{cho2014learning,hochreiter1997long,cho2014properties}.
%and modify the \emph{cell} to model spatial relationship and missing entries' pattern. 
However, due to the constraint of sequential computation over time, the training of RNN cannot be parallelized and thus is usually time-consuming. Moreover, the relationship between each two distant time stamps cannot be directly modeled.
%the imputation performance is limited by the order and length of the sequence while the measurements correlation is still not clearly interpreted.
%non-consecutive data points Phoenix 
Recently, the self-attention mechanism as shown in Fig.~\ref{Fig:Motivation}(b) has been proposed by the seminal work of \textit{Transformer}~\cite{vaswani2017attention} to get rid of the limitation of sequential processing, accelerating the training time substantially and improving the performance significantly on seq-to-seq tasks in Natural Language Processing (NLP) because the relevance between each two time stamps is captured explicitly.

In this paper, we are the first to adapt the self-attention mechanism to impute missing data in multivariate time series, which cover multiple geo-locations and contain multiple measurements as shown in Fig.~\ref{Fig:Motivation}(a).
%Since our time series data are geo-tagged and contain multiple measurements, we investigate several choices of modeling the attention and interaction in such unique multi-dimensional data.
In order to impute a missing value in such unique multi-dimensional data, it is very useful to look into available data in different dimensions (i.e. \textbf{time}, \textbf{location} and \textbf{measurement}), as shown in Fig.~\ref{Fig:Motivation}(c).
To this end, we investigate several choices of modeling self-attention across different dimensions.
In particular, we propose a novel Cross-Dimensional Self-Attention (CDSA) mechanism to capture the attention crossing all dimensions jointly yet in a decomposed, computationally inexpensive manner.
In summary, we make the following contributions:
% Inspired by the \textcolor{red}{success of self-attention mechanism}
% Internally, . 
% learn the association between words of two languages in a both respective and conjunctive manner.
%filling the missing values for time series of multiple measurements in a distributed network and requiring no assumption.  is calculated respectively while flexible attention mask is designed for various tasks. 

\begin{itemize}
    \item [(i)] We are the first to apply the self-attention mechanism to the multivariate, geo-tagged time series data imputation task, replacing the conventional RNN-based models to speed up training and directly model the relationship between each two data values in the input data.
    %Internally, the measurement correlation and spatial-temporal relationship is learned jointly in a distinguishable/decomposed manner.
    % the attention of different dimension, \ie, the temporal, spatial and measurement correlation is mathematically interpretable
    \item [(ii)] For the unique time series data of multiple dimensions (i.e. \textbf{time}, \textbf{location}, \textbf{measurement}), we comprehensively study several choices of modeling self-attention crossing different dimensions.
    Our CDSA mechanism models self-attention crossing all dimensions jointly yet in a dimension-wise decomposed way.
    We show that CDSA is computationally efficient and independent with the order of processing each dimension. 
    %\item [(iii)] This model directly treats the missed values as variable and learns without limitation bounded by time sequence. Furthermore, the back propagation from all possible dimension can be contributed to the missing value updating effectively.
    \item [(iii)] We extensively evaluate on three standard benchmarks and our newly collected traffic dataset. Experimental results show that our model outperforms the state-of-the-art models for both data imputation and forecasting tasks.
\end{itemize}

%%%%%%%%%%%%%%%%%%%%%%%%%%%%%%%%%%%%%%%%%%%%%%%%%%%%%%%%%%%%%%%%%%%%%%%%%%%%%%%%%%
%%%%%%%%%%%%%%==================   Related Work    ==================%%%%%%%%%%%%%
%%%%%%%%%%%%%%%%%%%%%%%%%%%%%%%%%%%%%%%%%%%%%%%%%%%%%%%%%%%%%%%%%%%%%%%%%%%%%%%%%%
\section{Related Work}

%Existing data imputation methods consist of statistical methods and deep learning methods.
\textbf{Statistical data imputation methods.}
Statistical methods \cite{ansley1984estimation,zhang2003time,shumway1982approach,nelwamondo2007missing,buuren2010mice} often impose assumptions over data and reconstruct the missed value by fitting a \textit{smooth curve} to the available values.
For instance, Kriging variogram model~\cite{stein2012interpolation} was proposed to capture the variance in data w.r.t. the geodesic distance.
%\eg, ARIMA, SARIMA~\cite{zhang2003time}, EM algorithm~\cite{nelwamondo2007missing} and MICE\cite{buuren2010mice} 
Matrix completion methods~\cite{acuna2004treatment,yu2016temporal,friedman2001elements,cai2010singular,ji2009accelerated,ma2011fixed} usually enforce low-rank constraint.
%By modeling the data as a distribution, \cite{yu2012solving,guerrero2008image} were proposed from a probabilistic perspective.
%mean/median filling~\cite{acuna2004treatment}, MF~\cite{yu2016temporal}, KNN~\cite{friedman2001elements},  SVD~\cite{cai2010singular}, Norm-minimization~\cite{ji2009accelerated}, Rank minimization~\cite{ma2011fixed},Kriging~\cite{stein2012interpolation}
%tries to fit the semi-variogram and distance into a parametrized model. Since multiple stations are located arbitrarily, graph is intuitively introduced to represent such spatial structure.
%However, the performance is always limited in the strong assumption \phoenix{and the available observation used for estimation is limited in a specific dimension.}

\textbf{RNN-based data imputation methods.} 
~\citet{li2018dcrnn_traffic} proposed DCGRU for seq-to-seq by adopting graph convolution~\cite{chung1997spectral,shi2009kalman,shuman2012emerging} to model spatio-temporal relationship.
%However, the graph requires designated spatial information and previous knowledge~\cite{shuman2012emerging} to keep its sparsity. 
\citet{luo2018multivariate} built GRUI by incorporating RNN into a Generative Adversarial Network (GAN).
% where each sensor.
%, labeled by various location and measurement, is indiscriminately viewed as a variable so that the time series of each variable is considered jointly.
%Nevertheless, the spatiotemporal and measurements correlation are mixed and indistinguishable.
\citet{cao2018brits} utilized bi-directional RNN and treated the missing values as trainable variables.
%so that the mediate back propagation from loss of available observation can contribute to the missing value updating.
Nevertheless, these RNN-based models fundamentally suffer from the constraint of sequential processing, which leads to long training time and prohibits the direct modeling of the relationship between two distant data values.

\textbf{Self-attention.}
%self-attention mechanism has been a versatile part of sequence dependencies modeling for eliminating the bound by distance in the time stamps~\cite{hochreiter2001gradient} and has been successfully applied on various tasks~\cite{cheng2016long,parikh2016decomposable,paulus2017deep,lin2017structured}. 
Recently, \citet{vaswani2017attention} introduced the \textit{Transformer} framework which entirely rely on self-attention, learning the association between each two words in a sentence. 
Then self-attention has been widely applied in seq-to-seq tasks such as machine translation, image generation~\cite{yang2016stacked,zhang2018self} and graph-structured data~\cite{velivckovic2017graph}.
% and formulating relationship in video sequence~\cite{wang2018non}. 
In this paper, we are the first to apply self-attention for multi-dimensional data imputation and specifically we investigate several choices of modeling self-attention crossing different data dimensions.

%algorithm to learn the interaction both inside and outside multiply dimension. 

%On the other hand, the correlation between different measurements can also be utilized. As shown in Figure \ref{Fig:Measurement-Relationship}, higher PM2.5 value often indicates higher concentration of PM10 while higher NO$_2$ usually forebode lower O$_3$. Thus, missing value of PM2.5 can be estimated from PM10 if applicable. 

%% file: Paper_Model.tex
% Uncertainty Phoenix
% transitional -- conventional (recommended)
% Section 3.2.1--3.2.3
%%%%%%%%%%%%%%%%%%%%%%%%%%%%%%%%%%%%%%%%%%%%%%%%%%%%%%%%%%%%%%%%%%%%%%%%%%%%%%%%%%
%%%%%%%%%%%%%%================   Model Description   ================%%%%%%%%%%%%%
%%%%%%%%%%%%%%%%%%%%%%%%%%%%%%%%%%%%%%%%%%%%%%%%%%%%%%%%%%%%%%%%%%%%%%%%%%%%%%%%%%
\section{Approach}\label{Sec:CDSA Mechanism}

\begin{figure}[t!]
    \centering
    \subfigure[Independent]
    {\label{Fig:Independent}\includegraphics[width=0.3665\linewidth]{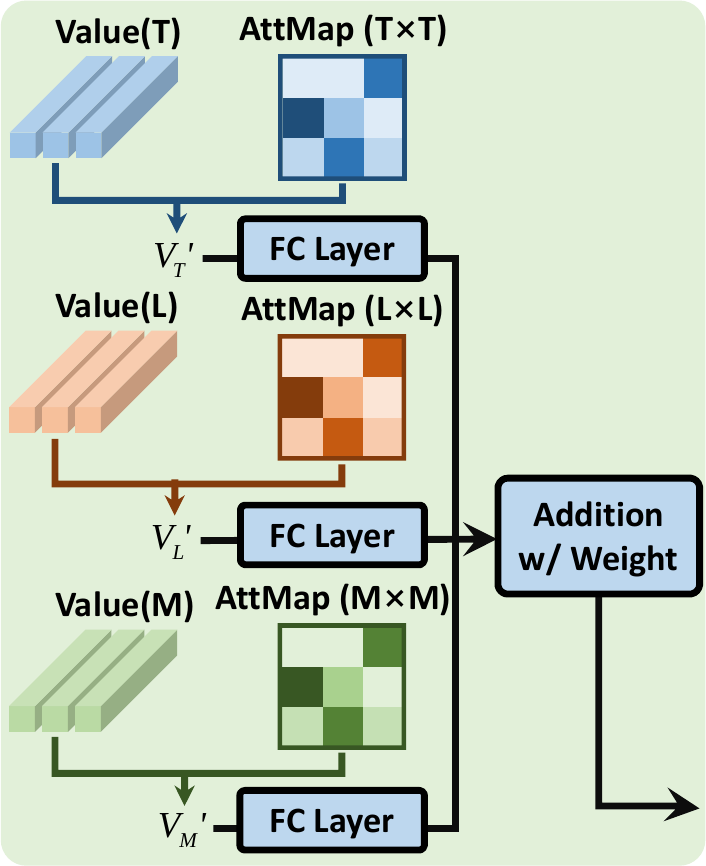}}
    \subfigure[Joint]
    {\label{Fig:Joint}\includegraphics[width=0.289\linewidth]{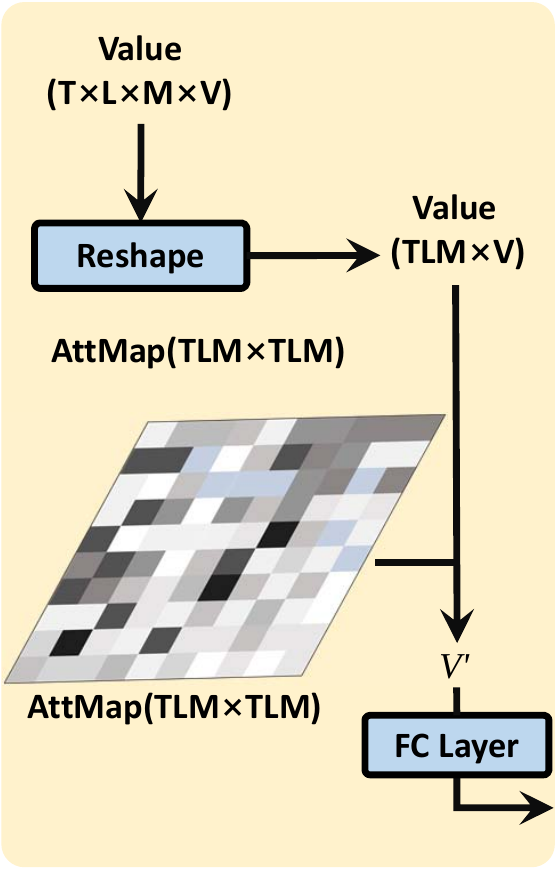}}
    \subfigure[Decomposed]
    {\label{Fig:Decomposed}\includegraphics[width=0.33\linewidth]{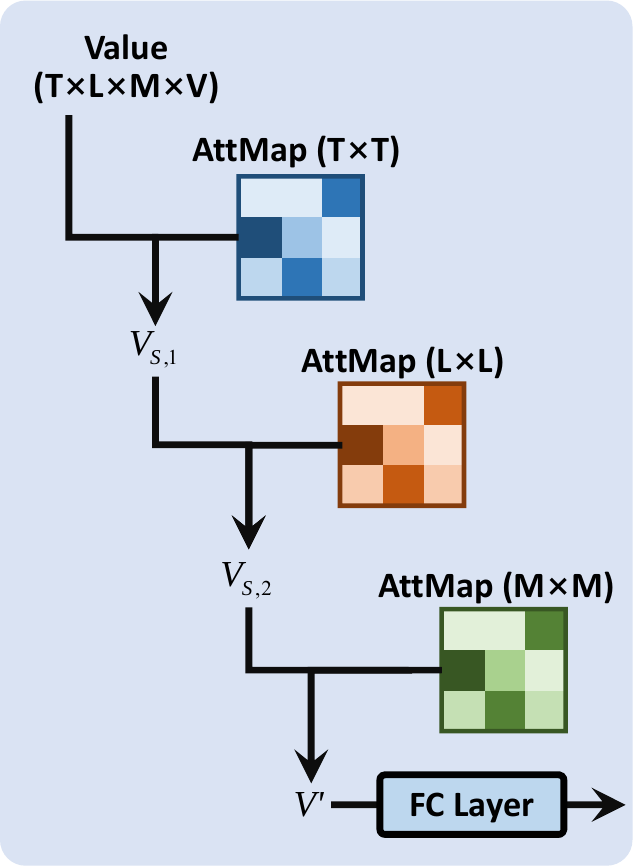}}
     \caption{Three choices of implementing our Cross-Dimensional Self-Attention mechanism}\label{Fig:Mechanism}
\end{figure}

In Sec.~\ref{Sec:Review}, we first review the conventional self-attention mechanism in NLP. 
In Sec.~\ref{Sec:CDSA Mode}, we propose three methods for computing attention map cross different dimensions.
In Sec.~\ref{Sec:Designed for Imputation}~and~\ref{Sec:Implementation}, we present details of using CDSA for missing data imputation.

\subsection{Conventional Self-Attention}\label{Sec:Review}

As shown in Fig.~\ref{Fig:Motivation}(b), for language translation task in NLP, given an input sentence, each word $\bm{x}_i$ is mapped into a \emph{Query} vector $\bm{q}_i$ of $d$-dim, a \emph{Key} vector $\bm{k}_i$ of $d$-dim, and a \emph{Value} vector $\bm{v}_i$ of $v$-dim.
The attention from word $\bm{x}_j$ to word $\bm{x}_i$ is effectively the scaled dot-product of $\bm{q}_i$ and $\bm{k}_j$ after Softmax, which is defined as
$\bm{A}(i,j) = \text{exp}(\bm{S}(i,j)) \big( \sum_{j=1}^T \text{exp}(\bm{S}(q,j)) \big)^{-1}$ where $\bm{S}(i,j) = \bm{q}_i \bm{k}_j \ts / \sqrt{d}$. 
Then, $\bm{v}_i$ is updated to $\bm{v}_i'$ as a weighted sum of all the \emph{Value} vectors, defined as $\bm{v}_i' = \sum_{j=1}^T \bm{A}(i,j)\bm{v}_j$, 
after which each $\bm{v}_i'$ is mapped to the layer output $\bm{x}_i'$ of the same size as $\bm{x}_i$.
In order to adapt the self-attention from NLP to our multivariate, geo-tagged time series data, a straightforward way is to view all data in a time stamp as one word embedding and model the self-attention over time.

%and $1 \leq j \leq T$
%\emph{Softmax} is applied on the Scores in the same each row to draw out $\bm{A}(i,j)$ so that the sum of units in each row is equal to 1. Thus, the dependency between different words are learned without the limitation of distance.
%\noindent where $\bm{S}(i,j) = \frac{\bm{q}_i \bm{k}_j \ts}{\sqrt{d}},~~0<j<T$
%
%\zs{combine notation here}
%the \textit{Scaled Dot-Product Attention} in~\cite{vaswani2017attention} is essentially a weighted sum .
% where the weight is calculated based on the model itself. 
%In each self-attention module, \textit{query, key} and \textit{value} are mapped respectively from embedded input through matrix multiplication.
%The attention map is basically calculated from dot multiplication between \textit{query} and \textit{key} where.
%Multi-head attention is applied further to learn the relationship between different words thoroughly. 
%
%In NLP scenario, the order of words can be viewed as a time sequence while the attention map is subsequently built in time. %By generalizing the idea of attention map from \emph{time} to multiple dimensions, we put forward our CDSA mechanism and study 3 different methods.

\subsection{Cross-Dimensional Self-Attention}\label{Sec:CDSA Mode}

In order to model Cross-Dimensional Self-Attention (CDSA), in this section we propose three solutions: (1) model attention within each dimension \textbf{independently} and perform late fusion; (2) model attention crossing all dimensions \textbf{jointly}; (3) model attention crossing all dimensions in a joint yet \textbf{decomposed} manner.
We assume the input data $\mathcal{X} \in \mathbb{R}^{T \times L \times M}$ has three dimensions corresponding time, location, measurement. 
$\mathcal{X}$ can be reshaped into two-dimensional matrices (\ie~ $\bm{X}_{\mathcal{T}} \in \mathbb{R}^{T \times LM}$, $\bm{X}_{\mathcal{L}} \in \mathbb{R}^{L \times MT}$, $\bm{X}_{\mathcal{M}} \in \mathbb{R}^{M \times TL}$) or an one-dimensional vector (\ie~$\bm{X} \in \mathbb{R}^{TLM \times 1}$). 
Similarly, this superscript may be applied on the \emph{Query}, \emph{Key} and \emph{Value}, \eg, $\mathcal{Q} \in \mathbb{R}^{T \times L \times M \times d}$, $\bm{Q}_{\mathcal{L}} \in \mathbb{R}^{L \times MTd}$ and $\bm{Q} \in \mathbb{R}^{TLM \times d}$.

%\az{\sout{Obviously, the above straightforward method based on conventional self-attention ignores the data relationship crossing different dimensions.}}
%In order to model Cross-Dimensional Self-Attention (CDSA) \az{\sout{} adapt the conventional self-attention mechanism to higher-dimensional data}

%Similarly, \emph{Query}, Key, and Value matrices will be expanded to 3D tensors.
%, termed as \emph{time} ($T$), \emph{location} ($L$), \emph{measurement} ($M$). 
%\az{\sout{$\mathcal{X}$ can be reshaped into a two-dimensional matrix %$\bm{X}^Z$ where the superscript indicates the kept dimension, \ie, 
%.
%Similarly, this superscript will also be applied on \textit{value} vectors.
%Similar notations will also be applied for reshaping Value vectors.} Similarly, Query, Key, and Value matrices will be expanded to 3D tensors. We define the reshape operator $\phi_T: \mathbb{R}^{T \times L \times M} \rightarrow \mathbb{R}^{T \times LM}$ which rearranges the values of an input tensor, e.g. $\mathcal{X}$ or Value, into a matrix. Similarly, we define $\phi_L: \mathbb{R}^{T \times L \times M} \rightarrow \mathbb{R}^{L \times MT}$ and $\phi_M: \mathbb{R}^{T \times L \times M} \rightarrow \mathbb{R}^{M \times TL}$.} 

%As shown in Fig. \ref{Fig:Mechanism}, three methods of our CDSA mechanism are provided while the performance comparison are described in Section \ref{Sec:Ablation}. Typically, we apply the \emph{Method 3} in the framework and achieve the best performance.

\subsubsection{Independent}\label{Sec:Independent}

As shown in Fig.~\ref{Fig:Independent}, the input $\mathcal{X}$ is reshaped into three input matrices $\bm{X}_{\mathcal{T}}$, $\bm{X}_{\mathcal{L}}$ and $\bm{X}_{\mathcal{M}}$. Three streams of self-attention layers are built to process each input matrix in parallel.
Such as the first layer in stream on $\bm{X}_{\mathcal{L}}$, each vector $\bm{X}_{\mathcal{L}}(l,:)$ of $MT$-dim is viewed as a word vector in NLP.
Following the steps in Sec.~\ref{Sec:Review}, $\bm{X}_{\mathcal{L}}(l,:)$ is mapped to $\bm{Q}_L(l,:)$ and $\bm{K}_L(l,:)$ of $d_L$-dim, as well as $\bm{V}_L(l,:)$ of $v_L$-dim.
The output of every stream's last layer are fused through element-wise addition, $\mathcal{X}' = \alpha_T\mathcal{X}_T' + \alpha_L\mathcal{X}_L' + \alpha_M\mathcal{X}_M'$, where the weights $\alpha_T$, $\alpha_L$ and $\alpha_M$ are trainable parameters. 
Besides, the hyper-parameters for each stream such as the number of layers, are set separately.

\subsubsection{Joint}\label{Sec:Joint}

%As shown in Fig.~\ref{Fig:Joint}, the three-dimensional input $\mathcal{X}$ is reshaped as a vector $\bm{x}$ and we adopt the traditional self-attention mechanism to directly model the cross-dimensional relationship as in Sec.~\ref{Sec:Review} and draw out the attention map $\bm{A}_{Joint}$ of dimension $TLW \times TLW$. 
%Zheng's Original Modification (next 2 lines)
As shown in Fig.~\ref{Fig:Joint}, the three-dimensional input $\mathcal{X}$ is reshaped as to $\bm{X}$. Each unit $\bm{X}(p)$ is mapped to $\bm{Q}(p,:)$ and $\bm{K}(p,:)$ of $d$-dim as well as $\bm{V}(p,:)$ of $v$-dim, where $p=p(t,l,m)$ denotes the index mapping from the 3-D cube to the vector form. In this way, an attention map of dimension $TLM \times TLM$ is built to directly model the cross-dimensional interconnection.

\subsubsection{Decomposed}\label{Sec:Decomposed}

The \emph{Independent} manner sets multiple attention sub-layers in each stream to model the dimension-specific attention but fail in modeling cross-dimensional dependency. 
In contrast, the \emph{Joint} manner learns the cross-dimensional relationship between units directly but results in huge computation workload.
To capture both the dimension-specific attention and cross-dimensional attention in a distinguishable way, we propose a novel \emph{Decomposed} manner.

As shown in Fig.~\ref{Fig:Decomposed}, the input $\mathcal{X}$ is reshaped as input matrices $\bm{X}_{\mathcal{T}}$, $\bm{X}_{\mathcal{L}}$, $\bm{X}_{\mathcal{M}}$ and $\bm{X}$. 
Each unit $\bm{X}(p)$ is mapped to vector $\bm{V}(p,:)$ of $v$-dim as in the \emph{Joint} while $\bm{X}_{\mathcal{T}}$, $\bm{X}_{\mathcal{L}}$ and $\bm{X}_{\mathcal{M}}$ are used for building attention map $\bm{A}_T$, $\bm{A}_L$,$\bm{A}_M$ individually as in the \emph{Independent}. 
The attention maps are applied on \emph{Value} vector in order as, 
\begin{equation}
    \bm{V}' = \bm{A} \bm{V} = \widetilde{\bm{A}}_{M}\bm{V}_{\text{S},2} = \widetilde{\bm{A}}_{M}\widetilde{\bm{A}}_{L}\bm{V}_{\text{S},1} = \widetilde{\bm{A}}_{M} \widetilde{\bm{A}}_{L} \widetilde{\bm{A}}_{T} \bm{V}.
    \label{Eq:Order in Figure}
\end{equation}
\noindent The attention map with~~$\widetilde{}$~~is reshaped from the original attention map and consistent with the calculation in \eqref{Eq:Order in Figure}, \eg, $\widetilde{\bm{A}}_{T} \in \mathbb{R}^{TLM \times TLM}$ is reshaped from $\bm{A}_{T} \in \mathbb{R}^{T \times T}$. More specifically,
\begin{equation}
    \begin{split}
        \widetilde{\bm{A}}_{T} &= \bm{A}_{T} \otimes \bm{I}_{L} \otimes \bm{I}_{M},\\
        \widetilde{\bm{A}}_{L} &= \bm{I}_{T} \otimes \bm{A}_{L} \otimes \bm{I}_{M},\\
        \widetilde{\bm{A}}_{M} &= \bm{I}_{T} \otimes \bm{I}_{L} \otimes \bm{A}_{M},
    \end{split}\label{Eq:Attention Map Reshape}
\end{equation}
\noindent where $\otimes$ denotes \textit{tensor product} and $\bm{I}$ is the \textbf{\textit{Identity}} matrix where the subscript indicates the size, \eg, $\bm{I}_T \in \mathbb{R}^{T \times T}$. 
Although the three reshaped attention maps are applied with a certain order, according to \eqref{Eq:Attention Map Reshape}, we show that each unit in $\bm{A}$ is effectively calculated as 
\begin{equation}\label{Eq:Attention Coefficient}
    \bm{A}(p_0,p_1) = \bm{A}_T(t_0,t_1) \bm{A}_L(l_0,l_1) \bm{A}_M(m_0,m_1),
\end{equation}
\noindent where $p_0 = p(t_0,l_0,m_0), p_1 = p(t_1,l_1,m_1)$.
Following the associativity of tensor product, we demonstrate
\begin{equation}\label{Eq:Decomposed Vector From}
    \widetilde{\bm{A}}_{\sigma(L)} \widetilde{\bm{A}}_{\sigma(M)} \widetilde{\bm{A}}_{\sigma(T)} = \bm{A}_{T} \otimes \bm{A}_{L} \otimes \bm{A}_{M},
\end{equation}
\noindent where $\sigma = \sigma$(T,L,M) denotes the arbitrary arrangement of sequence (T,L,M), \eg, (T,M,L). 
Effectively, the arrangement $\sigma$ is the order of attention maps to update $\bm{V}$. %\ie, $\bm{V}_{S,1} = \widetilde{\bm{A}}_{\sigma(T)}\bm{V}$, $\bm{V}_{S,2} = \widetilde{\bm{A}}_{\sigma(L)}\bm{V}_{S,1}$ and $\bm{V}' = \widetilde{\bm{A}}_{\sigma(M)}\bm{V}_{S,2}$.
As \eqref{Eq:Attention Coefficient}-\eqref{Eq:Decomposed Vector From} shows that the weight in $\bm{A}$ is decomposed as the product of weights in three dimension-specific attention maps,
the output and gradient back propagation are order-independent. Furthermore, we show in Supp that the cross-dimensional attention map has the following property:
\begin{equation}\label{Eq:Sum Property}
    \sum_{p_1=1}^{TLM}\bm{A}(p_0,p_1) = \sum_{t_1=1}^{T}\sum_{l_1=1}^{L}\sum_{m_1=1}^{M}\bm{A}_T(t_0,t_1) \bm{A}_L(l_0,l_1) \bm{A}_M(m_0,m_1) = 1.
\end{equation}
\noindent In summary, the \emph{Independent} builds attention stream for each dimension while the \emph{Joint} directly model the attention map among all the units. 
Our proposed CDSA is based on the \emph{Decomposed}, which forms a cross-dimensional attention map, out of three dimension-specific maps. 
As an alternative of the \emph{Decomposed}, the \emph{Shared} maps unit $\bm{X}(p)$ to $\bm{Q}(p,:)$ and $\bm{K}(p,:)$ of $d$-dim and calculates all three dimension-specific attention map, \eg, $\bm{A}_L = \text{Softmax}(\bm{Q}_{(L)} \bm{K}_{(L)} \ts / \sqrt{MTd})$.
%\az{We have two variants of CDSA, one which shares the same Query and Key to compute all three dimension-specific attention maps (\emph{Shared}(Q,K)), and another that computes dimension-specific Query and Key matrices using unshared parameters (\emph{Decomposed}(Q,K)). \sout{Each map can be drawn from either the same $\bm{Q}~\&~\bm{K}$ (\emph{Decomposed}(Q,K,V)) or different $\bm{Q}~\&~\bm{K}$ (\emph{Decomposed}(V)). }}
%\footnote{https://www.tensorflow.org/api\_docs/python/tf/profiler/profile/}
As shown in Table~\ref{Table:Computation Complexity}, by using Tensorflow \textit{profile} and fixing the hyper-parameters with detailed explanations in Supp., the \emph{Decomposed} significantly decreases the FLoating point OPerations (FLOPs) compared to the \emph{Joint} and requires less variables than the \emph{Independent}. Detailed comparisons are reported in Sec.~\ref{Sec:Discussions}. 
%(\ie~efficiency and accuracy) 
\begin{table}[H]
    \centering
    \caption{Computational complexity of several methods to implement CDSA}
    \begin{tabular}{c | c | c | c | c }
    \hline
    Methods &  \emph{Independent} & \emph{Joint} & \emph{Shared} & \emph{Decomposed} \\
    \hline
    FLOPs($\times 10^9$) & 0.39 & 3.22 & 0.21 & 0.24\\
    Number of Variables ($\times 10^5$) & 18.15 & 0.44 & 0.44 & 16.09\\
    \hline
    \end{tabular}
    \label{Table:Computation Complexity}
\end{table}

%correlation of between different dimensional are realized through mapping all the three attention maps on the shared \textit{value} vectors. 
%It is worth noticing that either the shared \textit{query and key} or different \textit{queries and keys} can be used for for different attention map calculation. We found the latter always lead to better result and the comparison will be explained in Section \ref{Sec:Ablation}.

%Similar to the associativity of matrix multiplication, 
%where the order of pre-multiplication and post-multiplication does not effect the final result, \ie, $\bm{A}\bm{M}\bm{B} = (\bm{A}\bm{M})\bm{B} = \bm{A}(\bm{M}\bm{B})$, 
%both forward inference and backward propagation is independent of the order.
%where the updating of value vector can be described as 

%%%%%%%%%%%%%%%%%%%%%%%%%%%%%%%%%%%%%%%%%%%%%%%%%%%%%%%%%%%%%%%%%%%%%%%%%%%%%%%%%%
%%%%%%%%%%%%==============   Architecture Description   ===============%%%%%%%%%%%
%%%%%%%%%%%%%%%%%%%%%%%%%%%%%%%%%%%%%%%%%%%%%%%%%%%%%%%%%%%%%%%%%%%%%%%%%%%%%%%%%%
\subsection{Framework}\label{Sec:Designed for Imputation}

\begin{figure}
    \centering
    \includegraphics[width=\linewidth]{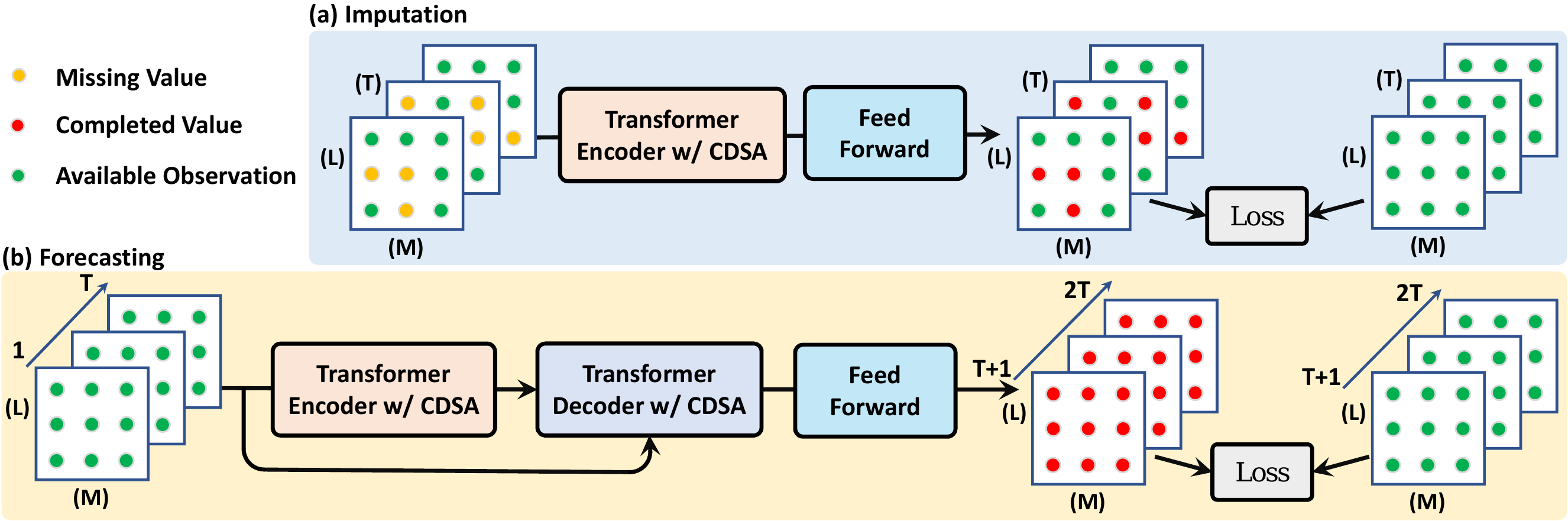}
    \caption{The framework employing CDSA for data imputation and forecasting.}
    \label{Fig:Archi_Impu}
\end{figure}

\textbf{Imputation}: As shown in Fig.~\ref{Fig:Archi_Impu}(a), we apply our CDSA mechanism in a \textit{Transformer \textbf{Encoder}}, a stack of $N=8$ identical layers with residual connection~\cite{he2016deep} and normalization~\cite{lei2016layer} as employed by~\cite{vaswani2017attention}. To reconstruct the missing (along with other) values of the input, we apply a fully connected \textit{Feed Forward} network on the final Value tensor, which is trained jointly with the rest of the model. 
%by sliding it on all three dimensions, and applying it to each scalar value:
%\begin{equation}\label{Eq:Reconstruction}
%    \hat{\mathcal{X}}[t, l, m] = f_{W_r}(V'[t, l, m])
%,\end{equation}
%where $f_{W_f}$ consists of 3 fully connected layers parametrized by $W_r$, and is .
%\az{\sout{Then, the \textbf{\textit{Encoder}} is followed by a fully connected \textit{Feed Forward} structure which is applied on each unit of the encoder output identically and individually.} }

%As the entry with missed value cannot be directly used for attention calculation with residual structure, inspired by~\cite{cao2018brits}, we first use a CDSA layer to derive a \textit{complement} input,
%\begin{gather}
%    \bm{M}(p(t,l,m)) = 
%    \begin{cases}
%        0 & \text{if} ~~ \bm{X}(p(t,l,m)) ~~ \text{is missed}\\
%        1 & \text{if} ~~ \bm{X}(p(t,l,m)) ~~ \text{is available}
%    \end{cases}\label{Eq:Mask}\\
%    \bm{X} \leftarrow \bm{X} \circ \bm{M} + \text{CDSA}(\bm{X}) \circ (1-\bm{M})
%\end{gather}
%\begin{equation}
%    \bm{X} \leftarrow \bm{X} \circ \bm{M} + \text{CDSA}(\bm{X}) \circ (1-\bm{M}),
%    \label{Eq:Complement Input} 
%\end{equation}
%\noindent where $\circ$ denotes element-wise multiplication and $\bm{M}$ is a binary matrix that uses $1$ to label the available observation and $0$ to label the missed value. Since the entry of missed value is originally set as 0 so that it has no impact for the calculation of its own complemented value. Then, the complemented $\bm{X}$ can be directly used for attention map calculation.

\textbf{Forecasting}: As shown in Fig.~\ref{Fig:Archi_Impu}(b), we apply our CDSA mechanism in \textit{Transformer} framework where we set $N=9$ for both encoder and decoder.
Similar to imputation, we use a fully connected feed forward network to generate the predicted values.
\subsection{Implementation Details}\label{Sec:Implementation}

We normalize each measurement of the input by subtracting the mean and dividing by standard deviation across training data. Then the entries with missed value are set 0. 
We use the Adam optimizer~\cite{kingma2014adam} to minimize the Root Mean Square Error (RMSE) between the prediction and ground truth. The model is trained on a single NVIDIA GTX 1080 Ti GPU. 
More details (\eg, network hyper-parameters, learning rate and batch size) can be found in Supp.

%\textbf{Version 1:} We apply \emph{Z-Score} normalization for each measurement respectively. Then the entries with missed value are set 0, which is essentially a mean-filling step.
%We use the Adam optimizer~\cite{kingma2014adam} to minimize the Root Mean Square Error (RMSE) between the prediction and ground truth tensors. The model is trained on a single NVIDIA GTX 1080 Ti GPU. 
%More details (\eg, network hyper-parameters, learning rate and batch size) can be found in Supp. \textbf{Version 2:} 

%% file: Paper_Result.tex
%%%%%%%%%%%%%%%%%%%%%%%%%%%%%%%%%%%%%%%%%%%%%%%%%%%%%%%%%%%%%%%%%%%%%%%%%%%%%%%%%%
%%%%%%%%%%%%%%================      Experiments      ================%%%%%%%%%%%%%
%%%%%%%%%%%%%%%%%%%%%%%%%%%%%%%%%%%%%%%%%%%%%%%%%%%%%%%%%%%%%%%%%%%%%%%%%%%%%%%%%%
\section{Experiments}\label{Sec:Experiments}

%Our proposed methods are applicable to a wide variety of applications. We mainly evaluate our CDSA in \emph{Decomposed} Manner on three standard benchmarks and our newly collected data. The rest two methods are used for ablation study and comparison.

\subsection{Datasets, Tasks, Evaluation Metrics}\label{Sec:Label&Task}

%\textbf{Version 1:} New York City Department of Transportation has set up various street cameras\footnote{https://nyctmc.org/}. Each camera takes a snapshot every few seconds. We recorded all the 186 cameras within Manhattan from 12/03/2015 to 12/26/2015.
%For each snapshot, we applied a Faster R-CNN model \cite{ren2015faster} to detect the cars and recorded the number of cars (\#car) in each snapshot. To aggregate such raw data into time series, for every non-overlapping 5-minute window, we averaged \#car to construct a $T \times 186 \times 1$ time series. We further split $T$ into 2-day intervals to create the final dataset, and split the intervals into training and testing sets.
%We evaluate our model for \emph{imputation} on the 1-month traffic volume dataset of 186 cameras in New York City Road Network.
\noindent \textbf{NYC-Traffic}. New York City Department of Transportation has set up various street cameras\footnote{https://nyctmc.org/}. 
Each camera keeps taking a snapshot every a few seconds.
The is collected around 1-month data from 12/03/2015 to 12/26/2015 for 186 cameras on Manhattan.
For each snapshot, we apply our trained faster-RCNN \cite{ren2015faster} vehicle detection model to detect the number of vehicles (\#vehicle) contained in each snapshot.
To aggregate such raw data into time series, for every non-overlapping 5-minute window, we averaged \#vehicle from each snapshot to obtain the average \#vehicle as the only measurement.
Finally, we obtained 186 time series where each value represents the average \#vehicle and the gap between two consecutive time stamps is 5 minutes.

The natural missing rate of the whole dataset is 8.43\%.
In order to simulate experiments for imputation, we further remove some entries and hold them as ground truth for evaluation.
The imputation task is to estimate values of these removed entries.
To mimic the natural data missing pattern, we model our manual removal as a \emph{Burst Loss}, which means at certain location the data is continuously missed for a certain period.
More details about vehicle detection and burst loss are be found in Supp.
To simulate various data missing extents, we vary the final missing rate after removal from 20\% to 90\%.
For each missing rate, we randomly select 432 consecutive time slots to train our model and evaluate the average RMSE of 5 trials.
The dataset will be released publicly.%. to facilitate future research.%\footnote{Link to be added}.

%follow \cite{luo2018multivariate}, we split this dataset for every 12 hours
%We model the generation of entries with missing value as , \ie, the missing data at different locations happen individually and the duration of missed value conforms with Gaussian Distribution.
%According to different missing rate, we set aside different available observation for evaluation. 
\noindent \textbf{KDD-2015}~\cite{zheng2015forecasting}. 
This dataset focuses on air quality and meteorology. It contains data recorded hourly, ending up with totally 8,759 time stamps.
PM2.5 measurement is recorded at 36 locations and Temperature and Humidity are recorded at 16 locations in Beijing from 05/01/2014 to 04/30/2015, with natural missing rate 13.3\%, 21.1\% and 28.9\% respectively. 
We treat those two subsets as two separate tasks and evaluate our method on each task separately.
Following \cite{yi2016st}, data in $3^{rd}, 6^{th}, 9^{th}$ and $12^{th}$ months are for testing and the remaining months are for training.
We randomly select 36 consecutive time slots to train our model and evaluate Mean Absolute Error (MAE) as well as Mean Relative Error (MRE).

In order to simulate experiments for imputation, besides the natural missing data, for PM2.5 we follow the strategy used in \cite{yi2016st,cao2018brits,zhou2018recover} to further manually remove entries and hold their values as ground truth.
The imputation task is to predict values of these manually removed entries.
For Temperature and Humidity, we follow  \cite{zhou2018recover} to randomly hold 20\% of available data.
%use 5 trials to compute the average MAE and MRE results.

\noindent \textbf{KDD-2018} ~\cite{KDDCup}.
Like \textbf{KDD-2015}, \textbf{KDD-2018} is an Air Quality and Meteorology dataset recorded hourly from 01/01/2017 to 04/30/2017. 
As indicated in \cite{luo2018multivariate}, 11 locations and 12 measurements are selected.
The natural missing rate is 6.83\%.
In order to simulate experiments for imputation, we follow \cite{luo2018multivariate} to split the data to every 48 hours, randomly hold values of some available entries and vary the missing rate from 20\% to 90\%.
Mean Squared Error (MSE) is used for evaluation.

\noindent \textbf{METR-LA} \cite{jagadish2014big}.
We follow \cite{li2018dcrnn_traffic} to use this dataset for traffic speed forecasting.
This dataset contains traffic speed at 207 locations recorded every 5 minutes for 4 months ranging from 03/01/2012 to 06/30/2012.
%he missing rate is \textcolor{gray}{8.73\%}.
Following \cite{li2018dcrnn_traffic}, 80\% of data at the beginning of these 4 months is used for training and the remaining 20\% is for testing.
In order to simulate the forecasting scenario, within either training or testing set, every time series of consecutive 2 hours are enumerated.
For each time series, data in the first hour is treated as input and data in the second hour is to be predicted.
We respectively evaluate the forecasting results at 15-th,
30-th, 60-th minutes in the second 1 hour and also evalaute the average evaluation results within the total 1 hour.
We use RMSE, MAE and Mean Absolute Percentage Error (MAPE) as evaluation metrics.
 %Also, the missed data is complemented with mean value and not counted in the evaluation. 

\subsection{Comparisons with State-of-the-art}

\noindent\textbf{NYC-Traffic}.
In Table~\ref{Table:NYC}, our CDSA consistently outperforms traditional methods including Auto Regressive, Kriging expo, Kriging linear and a recent RNN-based method (\ie~MTSI,~BRITS,~DCRNN) over a wide range of missing rate.
Because CDSA leverages the self-attention mechanism to avoid sequential processing of RNN and directly model the relationship between distant data.
Detailed overview of baselines can be found in Supp.

\noindent\textbf{KDD-2015}.
Table~\ref{Table:KDD-2015} shows that for PM2.5, our method outperforms the traditional methods significantly and achieves comparable MAE as IIN~\cite{zhou2018recover} while better MRE than IIN~\cite{zhou2018recover}.
For Temperature and Humidity, our method consistently outperforms state-of-the-art methods.
%the results in parentheses are the evaluation by training the model for each measurement individually and we can find that the value between different measurements contributes to the estimation of other measurement and improve the data imputation performance.

\begin{table}[!htb]
    \centering
    \caption{RMSE on dataset \textbf{NYC-Traffic} for comparisons with SOTA}
    \begin{tabular}{c | c c c c c c c c }
    %\begin{tabular}{c | c | c | c | c | c | c | c | c |}
        \hline
         Model $\backslash$ Missing Rate & 20\% & 30\% & 40\% & 50\% & 60\% & 70\% & 80\% & 90\% \\ 
         \hline\hline
         Auto Regressive & 2.354 & 2.357 & 2.359 & 2.362 & 2.364 & 2.652 & 2.796 & 3.272\\
         Kriging expo & 2.142 & 2.145 & 2.157 & 2.152 & 2.155 & 2.165 & 2.182 & 2.231\\
         Kriging linear & 2.036 & 2.008 & 2.031 & 2.038 & 2.056 & 2.074 & 2.111 & 2.194\\
         MTSI~\cite{luo2018multivariate} & 1.595 & 1.597 & 1.603 & 1.605 & 1.608 & 1.641 & 1.672 & 1.834\\
         BRITS~\cite{cao2018brits}  & 1.337 & 1.339 & 1.341 & 1.355 & 1.376 & 1.395 & 1.408 & 1.477\\
         DCRNN~\cite{li2018dcrnn_traffic} & 1.397 & 1.399 & 1.401 & 1.419 & 1.432 & 1.443 & 1.459 & 1.601\\ 
         \textbf{CDSA (ours)} & \textbf{1.203} & \textbf{1.208} & \textbf{1.211} & \textbf{1.214} & \textbf{1.215} & \textbf{1.217} & \textbf{1.234} & \textbf{1.377}\\
         \hline\hline
    \end{tabular}
    \label{Table:NYC}
\end{table}

\begin{table}[!htb]
    \centering
    \caption{MAE/MRE on dataset \textbf{KDD-2015} for comparisons with SOTA}
    \begin{tabular}{c | c c | c c | c c }
        \hline
         %\multirow{2}{*}{Model $\backslash$ Dataset} & \multicolumn{2}{c|}{Air Quality} & \multicolumn{4}{c|}{Meteorology} \\ 
         \multirow{2}{*}{Model $\backslash$ Dataset} & \multicolumn{2}{c|}{PM2.5} & \multicolumn{2}{c|}{TEMP} & \multicolumn{2}{c}{HUM}\\\cline{2-7}
         & MAE & MRE & MAE & MRE & MAE & MRE\\
         \hline
         Mean & 55.51 & 77.97\% & 9.21 & 97.56\% & 20.34 & 57.85\% \\
         KNN & 29.79 & 41.85\% & 1.26 & 19.83\% & 7.28 & 16.22\% \\
         MICE & 27.42 & 38.52\% & 1.23 & 18.29\% & 6.97 & 15.87\%\\
         ST-MVL~\cite{yi2016st} & 12.12 & 17.40\% & 0.68 & 4.59\% & 3.37 & 5.91\%\\
         MTSI~\cite{luo2018multivariate} & 13.34 & 18.01\% & 0.71 & 4.67\% & 3.51 & 6.21\% \\
         BRITS~\cite{cao2018brits} & 11.56 & 16.65\% & 0.63 & 4.16\% & 3.31 & 5.68\%\\
         DCRNN~\cite{li2018dcrnn_traffic} & 12.33 & 17.82\% & 0.69 & 4.59\% & 2.95 & 5.12\%\\
         IIN~\cite{zhou2018recover} & \textbf{10.63} & 15.31\% & 0.63 & 4.22\% & 2.90 & 5.09\%\\
         \textbf{CDSA (ours)} & 10.67 & \textbf{14.89}\% & \textbf{0.61} & \textbf{4.15}\% & \textbf{2.81} & \textbf{4.92}\%\\
        \hline\hline
    \end{tabular}
    \label{Table:KDD-2015}
\end{table}
%(\textcolor{gray}{0.67}) (\textcolor{gray}{4.41\%}) (\textcolor{gray}{3.02}) (\textcolor{gray}{5.34\%})

\begin{table}[!htb]
    \centering
    \caption{MSE on dataset \textbf{KDD-2018} for comparisons with SOTA}
    \begin{tabular}{c | c c c c c c c c }
    %\begin{tabular}{c | c | c | c | c | c | c | c | c |}
        \hline
         Model $\backslash$ Missing Rate & 20\% & 30\% & 40\% & 50\% & 60\% & 70\% & 80\% & 90\% \\ 
         \hline\hline
         Mean Filling & 0.916 & 0.907 & 0.914 & 0.923 & 0.973 & 0.935 & 0.937 & 1.002 \\
         KNN Filling & 0.892 & 0.803 & 0.776 & 0.798 & 0.856 & 0.852 & 0.873 & 1.243 \\
         MF Filling & 0.850 & 0.785 & 0.787 & 0.772 & 0.834 & 0.805 & 0.860 & 1.196 \\
         MTSI~\cite{luo2018multivariate} & 0.844 & 0.780 & 0.753 & 0.743 & 0.803 & 0.780 & 0.837 & 1.018 \\
         BRITS~\cite{cao2018brits}  & 0.455 & 0.421 & 0.372 & 0.409 & 0.440 & 0.482 & 0.648 & 0.725\\
         DCRNN~\cite{li2018dcrnn_traffic} & 0.579 & 0.565 & 0.449 & 0.506 & 0.589 & 0.622 & 0.720 & 0.861\\ \hline
         \textbf{CDSA (ours)} & \textbf{0.373} & \textbf{0.393} & \textbf{0.287} & \textbf{0.291} & \textbf{0.387} & \textbf{0.495} & \textbf{0.521} & \textbf{0.631} \\
        \hline\hline
    \end{tabular}
    \label{Table:KDD-2018}
\end{table}

\noindent\textbf{KDD-2018}. Table~\ref{Table:KDD-2018} shows that our proposed method again achieves significant improvements over the traditional methods and the RNN-based MTSI method which reported the best number on this dataset so far.%, all sensors, labeled by various location and measurement, are indiscriminately viewed as variables so that it fails to consider the spatio-temporal and measurements correlation in a distinguishable manner. In contrast, our CDSA mechanism in \emph{Decomposed} manner can solve this problem.

\noindent \textbf{METR-LA}. Table~\ref{Table:METR-LA-Horizon} shows that for the forecasting task, our CDSA method outperforms previous methods in most cases.
In particular, our method demonstrates clear improvement at long-term forecasting such as 60 min.
This again confirms the effectiveness of directly modeling the relationship between two distant data values using self-attention mechanism.

\begin{table}[!htb]
    \centering
    \caption{MAE/RMSE/MAPE on dataset \textbf{METR-LA} for comparisons with SOTA}
    \begin{tabular}{c | c c c | c c c }
        \hline 
        \multirow{2}{*}{Model} & \multicolumn{3}{c|}{15min} & \multicolumn{3}{c}{30min} \\\cline{2-7}
         & MAE & RMSE & MAPE & MAE & RMSE & MAPE \\
         \hline
         \text{FC-LSTM}~\cite{sutskever2014sequence} & 3.44 & 6.3 & 9.6\% & 3.77 & 7.23 & 10.9\% \\
         \text{MTSI}~\cite{luo2018multivariate} & 3.75 & 7.31 & 10.52\% & 3.89 & 7.73 & 11.04\%  \\
         \text{BRITS}~\cite{cao2018brits} & 2.86 & 5.46 & 7.49\% & 3.37 & 6.78 & 9.13\% \\
         \text{DCRNN}~\cite{li2018dcrnn_traffic} & 2.77 & 5.38 & 7.3\% & 3.15 & 6.45 & 8.8\% \\
         \text{DST-GCNN}~\cite{wang2018dynamic} & 2.68 & 5.35 & 7.2\% & \textbf{3.01} & 6.23 & 8.52\% \\
         \text{GaAN}~\cite{zhang2018gaan} & \textbf{2.71} & 5.24 & \textbf{6.99\%} & 3.12 & 6.36 & 8.56\% \\
         \textbf{CDSA(ours)} & 3.01 & \textbf{5.08} & 7.82\% & 3.14 & \textbf{5.38} & \textbf{8.30\%} \\
        \hline
        \multirow{2}{*}{Model} & \multicolumn{3}{c|}{60min} & \multicolumn{3}{c}{Mean} \\\cline{2-7}
        & MAE & RMSE & MAPE & MAE & RMSE & MAPE \\
         \hline
         \text{FC-LSTM}~\cite{sutskever2014sequence} & 4.37 & 6.89 & 13.2\% & 3.86 & 7.41 & 11.2\% \\
         \text{MTSI}~\cite{luo2018multivariate} & 4.22 & 8.39 & 12.15\% & 4.01 & 7.59 & 10.85\% \\
         \text{BRITS}~\cite{cao2018brits} & 3.65 & 7.66 & 10.55\% & 3.32 & 6.96 & 9.47\% \\
         \text{DCRNN}~\cite{li2018dcrnn_traffic} & 3.60 & 7.59 & 10.50\% & 3.28 & 6.80 & 8.87\% \\
         \text{DST-GCNN}~\cite{wang2018dynamic} & 3.41 & 7.47 & 10.25\% & - & - & - \\
         \text{GaAN}~\cite{zhang2018gaan} & 3.6 & 7.6 & 10.5\% & \textbf{3.16} & 6.41 & 8.72\% \\
         \textbf{CDSA(ours)} & \textbf{3.40} & \textbf{6.27} & \textbf{9.76\%} & \textbf{3.16} & \textbf{5.48} & \textbf{8.50\%}\\
        \hline\hline
    \end{tabular}
    \label{Table:METR-LA-Horizon}
\end{table}

\subsection{Discussions}\label{Sec:Discussions}
%\subsubsection{Ablation Study}\label{Sec:Ablation}

\textbf{The effects of different training losses}: For the forecasting task in \textbf{METR-LA}, we compare the performance by setting different training loss in Table \ref{Table:Performance Vs Loss} and we can see the performance with RMSE as loss metric achieves the best performance.

\begin{table}[H]
    \centering
    \caption{Comparisons of different losses in \emph{CDSA} on \textbf{METR-LA}}
    \begin{tabular}{c | c c | c c | c c | c c }
        \hline
        Time & 30min & Ave & 30min & Ave & 30min & Ave & 30min & Ave \\
        \hline
        Metric $\backslash$ Loss & \multicolumn{2}{c|}{RMSE} & \multicolumn{2}{c|}{MSE} & \multicolumn{2}{c|}{MAE} & \multicolumn{2}{c}{(RMSE+MAE)/2} \\ 
        \hline
        MAE & \textbf{3.14} & \textbf{3.16} & 3.43 & 3.41 & 3.28 & 3.33 & 3.21 & 3.25 \\
        RMSE & \textbf{5.38} & \textbf{5.48} & 6.20 & 6.11 & 5.67 & 5.83 & 5.55 & 5.70 \\
        MAPE & \textbf{8.30\%} & \textbf{8.50\%} & 9.32\% & 9.19 & 8.70\% & 9.00\% & 8.53\% & 8.80\% \\
        \hline\hline
    \end{tabular}
    \label{Table:Performance Vs Loss}
\end{table}

\textbf{Ablation study of different cross-dimensional self-attention manners}:
We compare the performance for different solutions in CDSA mechanism on the four datasets listed above.
%\\ \noindent \textbf{Efficiency}: 
The way of attention modeling determines the computational efficiency. As shown in Table \ref{Table:Computation Complexity}, 
since the \emph{Independent} calculate dimension-specific \emph{Value} vectors in parallel, the number of variables and FLOPs are larger than those of the \emph{Decomposed}.
As the \emph{Joint} and the \emph{Shared} all share the variables for each units, the number of variables is small and basically equal with each other. As the \emph{Joint} builds an huge attention map, its FLOPs is much larger than the rest.
Since the \emph{Decomposed} draws attention maps as the \emph{Independent} but shares \emph{Value} as the \emph{Joint}, it reduces the computational complexity significantly.
As shown in Table~\ref{Table:Methods-Ablation-NYC}, \ref{Table:Methods-Ablation-NYC}, \ref{Table:Methods-Ablation-NYC}, \ref{Table:Methods-Ablation-NYC}, we evaluate these methods on \textbf{NYC-Traffic}, \textbf{KDD-2015}, \textbf{KDD-2018} and \textbf{METR-LA} datasets and the \emph{Decomposed} always achieves the best performance.

\begin{table}[!htb]
    \centering
    \caption{Comparisons of different manners to implement CDSA on dataset \textbf{NYC-Traffic}.}
    \begin{tabular}{c | c c c c c c c c }
        \hline
        Model $\backslash$ Missing Rate & 20\% & 30\% & 40\% & 50\% & 60\% & 70\% & 80\% & 90\%\\
        \hline
        CDSA(Independent) & 1.327 & 1.327 & 1.331 & 1.355 & 1.362 & 1.379 & 1.393 & 1.425\\
        CDSA(Joint) & \multicolumn{8}{c}{Not Applicable due to memory usage limitation} \\
        CDSA(Shared) & 1.637 & 1.645 & 1.651 & 1.657 & 1.684 & 1.729 & 1.733 & 1.935 \\ \hline
        \textbf{CDSA(Decomposed)} &\textbf{1.204} & \textbf{1.208} & \textbf{1.211} & \textbf{1.214} & \textbf{1.215} & \textbf{1.217} & \textbf{1.235} & \textbf{1.377}\\
        \hline\hline
        %\multicolumn{2}{c}{Format: Mean (Variance$\times 10 ^ {-2}$)} & \multicolumn{3}{c}{  }\\
    \end{tabular}
    \label{Table:Methods-Ablation-NYC}
\end{table}

\begin{table}[!htb]
    \centering
    \caption{Comparisons of different manners to implement CDSA on dataset \textbf{KDD-2018}.}
    \begin{tabular}{c | c c c c c c c c }
        \hline
        Model $\backslash$ Missing Rate & 20\% & 30\% & 40\% & 50\% & 60\% & 70\% & 80\% & 90\%\\
        \hline
        CDSA(Independent) &0.482 & 0.523 & 0.351 & 0.366 & 0.484 & 0.573 & .608 & 0.721\\
        CDSA(Joint) & 0.451 & 0.497 & 0.317 & 0.336 & 0.404 & 0.520 & 0.558 & 0.677 \\
        CDSA(Shared) & 0.783 & 0.799 & 0.672 & 0.692 & 0.784 & 0.793 & 0.791 & 0.832 \\ \hline
        \textbf{CDSA(Decomposed)} & \textbf{0.373} & \textbf{0.393} & \textbf{0.287} & \textbf{0.291} & \textbf{0.387} & \textbf{0.495} & \textbf{0.521} & 0\textbf{.631}\\
        \hline\hline
    \end{tabular}
    \label{Table:Methods-Ablation-KDD-2018}
\end{table}

\begin{table}[!htb]
    \centering
    \caption{Comparisons of different manners to implement CDSA on dataset \textbf{KDD-2015}.}
    \begin{tabular}{c | c c | c c | c c }
        \hline
        \multirow{2}{*}{Model $\backslash$ Dataset} & \multicolumn{2}{c|}{PM2.5} & \multicolumn{2}{c|}{TEMP} & \multicolumn{2}{c}{HUM} \\\cline{2-7}
        & MAE & MRE & MAE & MRE & MAE & MRE \\
        \hline
        CDSA(Independent) & 11.54 & 16.01\% & 0.68 & 4.40\% & 3.19 & 5.42\%\\
        CDSA(Joint) & 11.20 & 15.52\% & 0.65 & 4.27\% & 3.05 & 5.37\% \\
        CDSA(Shared) & 13.85 & 19.26\% & 0.75 & 5.18\% & 3.56 & 6.47\% \\ \hline
        \textbf{CDSA(Decomposed)} & \textbf{10.67} & \textbf{14.89}\% & \textbf{0.61} & \textbf{4.15\%} & \textbf{2.81} & \textbf{4.92\%} \\
        \hline\hline
    \end{tabular}
    \label{Table:Methods-Ablation-KDD-2015}
\end{table}

\begin{table}[!htb]
    \centering
    \caption{Comparisons of different manners to implement CDSA on dataset \textbf{METR-LA}.}%General baselines comparison on 
    \begin{tabular}{c | c c c | c c c }
        \hline
        \multirow{2}{*}{Model $\backslash$ Dataset} & \multicolumn{3}{c|}{60 min} & \multicolumn{3}{c}{Mean}\\\cline{2-7}
        & MAE & RMSE & MAPE & MAE & RMSE & MAPE \\
        \hline
        CDSA(Independent) & 3.54 & 7.02 & 10.29\% & 3.25 & 6.29 & 8.81\% \\
        CDSA(Joint) & 3.63 & 7.62 & 10.54\% & 3.30 & 6.83 & 9.43\% \\
        CDSA(Shared) & 3.92 & 7.93 & 11.17\% & 3.53 & 7.33 & 10.26\% \\ \hline
        \textbf{CDSA(Decomposed)} & \textbf{3.40} & \textbf{6.27} & \textbf{9.76\%} & \textbf{3.16} & \textbf{5.48} & \textbf{8.50\%} \\
        \hline\hline
        %\multicolumn{2}{c}{Evaluated in one trial.}
    \end{tabular}
    \label{Table:Methods-Ablation-METR-LA}
\end{table}

\begin{figure}[!htb]
    \centering
    \subfigure[Attention Map]
    {\label{Fig:Visualization-AQ-KDD2018}\includegraphics[width=0.27\linewidth]{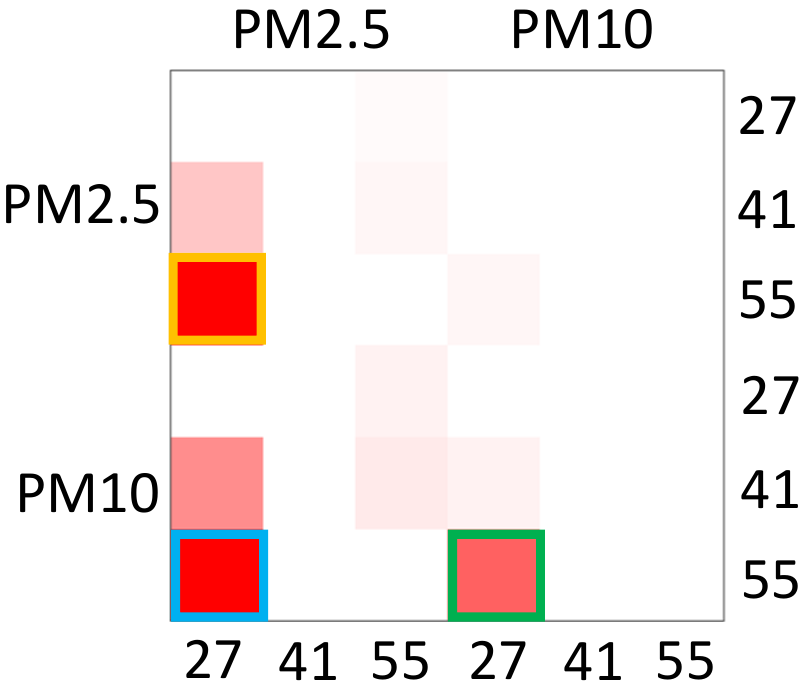}}\hfill
    \subfigure[Air Quality Time Series Data]
    {\label{Fig:Measurement-Relationship}\includegraphics[height=0.21\linewidth, width=0.67\linewidth]{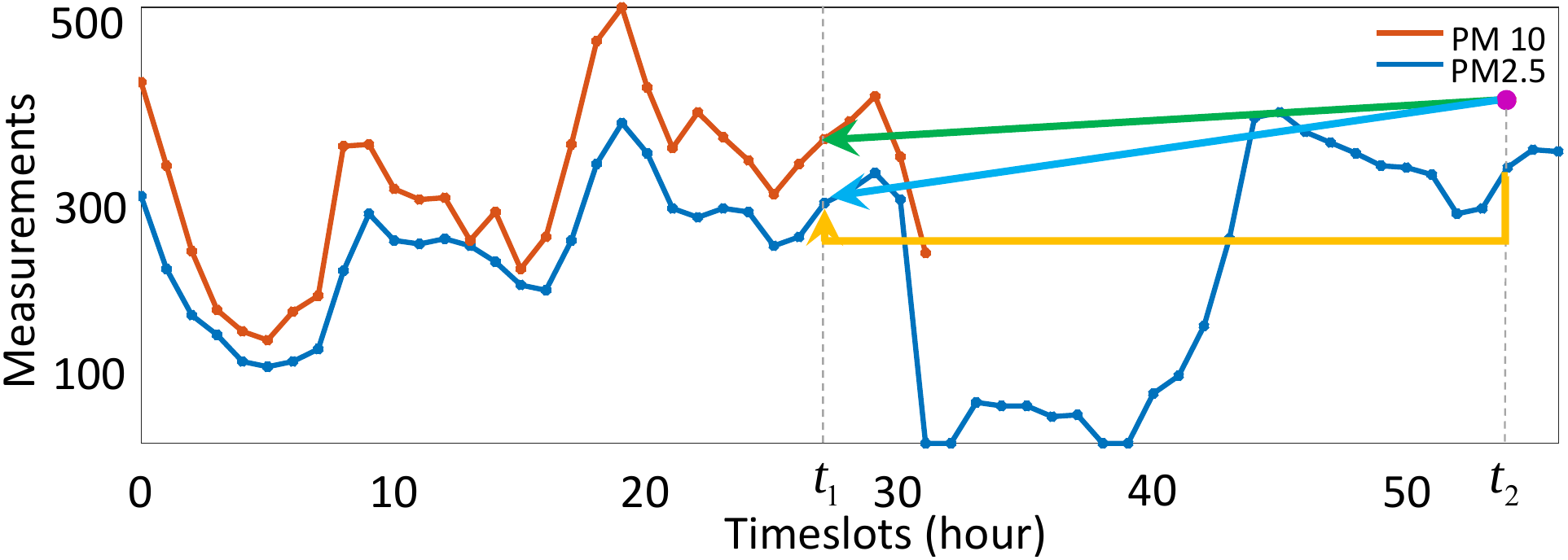}}
     \caption{Visualization of the cross-dimensional self-attention on \textbf{KDD-2015}. (a) Part of \textit{Time}-\textit{Measurement} attention map. (b) Two time series of PM2.5 and PM10. The value at \textcolor{purple}{purple} dot is missing and our model predicts its value based on other values. The arrow in (b) represents attention whose score is highlighted with bounding box in (a) of the same color.}
     \label{Fig:Visualiation-Confirm}
\end{figure}

\textbf{Attention Map Visualization}: %[to-be-edited]
Fig.~\ref{Fig:Visualiation-Confirm} shows an PM10 imputation example in location \textit{fangshan} at $t_2$. 
Since the pattern of PM2.5 around $t_2$ is similar to that at $t_1$, the attention in \textcolor{orange}{orange} box is high. As we can see that PM2.5 and PM10 are strongly correlated %($\rho_{\text{PM2.5, PM10}} = 0.8278$)
, in order to impute PM10 at $t_2$, our model utilizes PM10 at $t_1$ (\textcolor{green}{green} arrow) and PM2.5 at $t_1$ (\textcolor{blue}{blue} arrow), which crosses dimensions.

%\subsubsection{Visualization}\label{Sec:Visualization}

%Take the measurement at \textcolor{gray}{shunyi} in \textbf{KDD-2018} as an example, we select the time slots where PM2.5 and PM10, NO$_2$ and O$_3$ are both available respectively and calculate the correlation coefficient. The first 300 time slots as shown in Fig.~\ref{Fig:Measurement-Relationship} where PM2.5 and PM10 are significantly positive correlated (0.8172) and NO$_2$ and O$_3$ are significantly negative correlated (-0.5726). Part of the attention map in one head of the top CDSA sub-layer in \textbf{Encoder} is shown in Fig.~\ref{Fig:Visualization-AQ-KDD2018}. The square with deep red indicates a relatively higher relation and consistent with the measurement correlation shown in Fig.~\ref{Fig:Measurement-Relationship}.